\documentclass[10pt, a4paper]{article}
\usepackage{lrec2022} 
\usepackage{multibib}
\newcites{languageresource}{Language Resources}
\usepackage{graphicx}
\usepackage{tabularx}
\usepackage{soul}

\usepackage{lipsum}

\usepackage{lipsum}
\usepackage{xcolor}
\usepackage{hyperref}
\usepackage{placeins}
\usepackage{booktabs}

\usepackage{titlesec}
\titleformat{\section}{\normalfont\large\bfseries\center}{\thesection.}{1em}{}
\titleformat{\subsection}{\normalfont\SmallTitleFont\bfseries\raggedright}{\thesubsection.}{1em}{}
\titleformat{\subsubsection}{\normalfont\normalsize\bfseries\raggedright}{\thesubsubsection.}{1em}{}
\renewcommand\thesection{\arabic{section}}
\renewcommand\thesubsection{\thesection.\arabic{subsection}}
\renewcommand\thesubsubsection{\thesubsection.\arabic{subsubsection}}

\usepackage{epstopdf}
\usepackage[utf8]{inputenc}

\usepackage{hyperref}
\usepackage{xstring}

\usepackage{color}

\title{Extracting Space Situational Awareness Events from News Text}

\name{Zhengnan Xie$^{\dagger}$, Alice Saebom Kwak$^{\dagger}$, Enfa George$^{\dagger}$, Laura W. Dozal$^{\dagger}$, \\ {\bf \large Hoang Van$^{\dagger}$, Moriba Jah$^{\ddagger}$, Roberto Furfaro$^{\dagger}$, Peter Jansen$^{\dagger}$} }

\address{$^{\dagger}$University of Arizona, USA \\
         $^{\ddagger}$University of Texas at Austin, USA \\
         \texttt{pajansen@arizona.edu}}

\abstract{
Space situational awareness typically makes use of physical measurements from radar, telescopes, and other assets to monitor satellites and other spacecraft for operational, navigational, and defense purposes.  In this work we explore using textual input for the space situational awareness task.  We construct a corpus of 48.5k news articles spanning all known active satellites between 2009 and 2020.  Using a dependency-rule-based extraction system designed to target three high-impact events -- spacecraft launches, failures, and decommissionings, we identify 1,787 space-event sentences that are then annotated by humans with 15.9k labels for event slots.  We empirically demonstrate a state-of-the-art neural extraction system achieves an overall F1 between 53 and 91 per slot for event extraction in this low-resource, high-impact domain.
 \newline  }
\begin{document}

\maketitleabstract

\section{Introduction}

\noindent Space Situational Awareness (SSA) is the decision-making knowledge required to predict, avoid, operate through, or recover from the loss, disruption, or degradation of space services, capabilities, or activities \cite{cai2020possibilistic}. Most governments approach SSA as a physical measurement problem, using global arrays of optical telescopes, radar facilities, radio listening posts, and other assets to directly measure spacecraft properties, orbits, or communications.  In addition to being costly, this approach has several known limitations, chief among them that physical measurements (such as that a \textit{spacecraft is spinning}) alone cannot fully capture causal information (e.g. \textit{a spacecraft suffered a gyro failure}) that can be critical when interpreting and responding to events within the space object population \cite{walls2016assessing}.  

\begin{figure}[t!]
    \centering
    \includegraphics[width=2.5in]{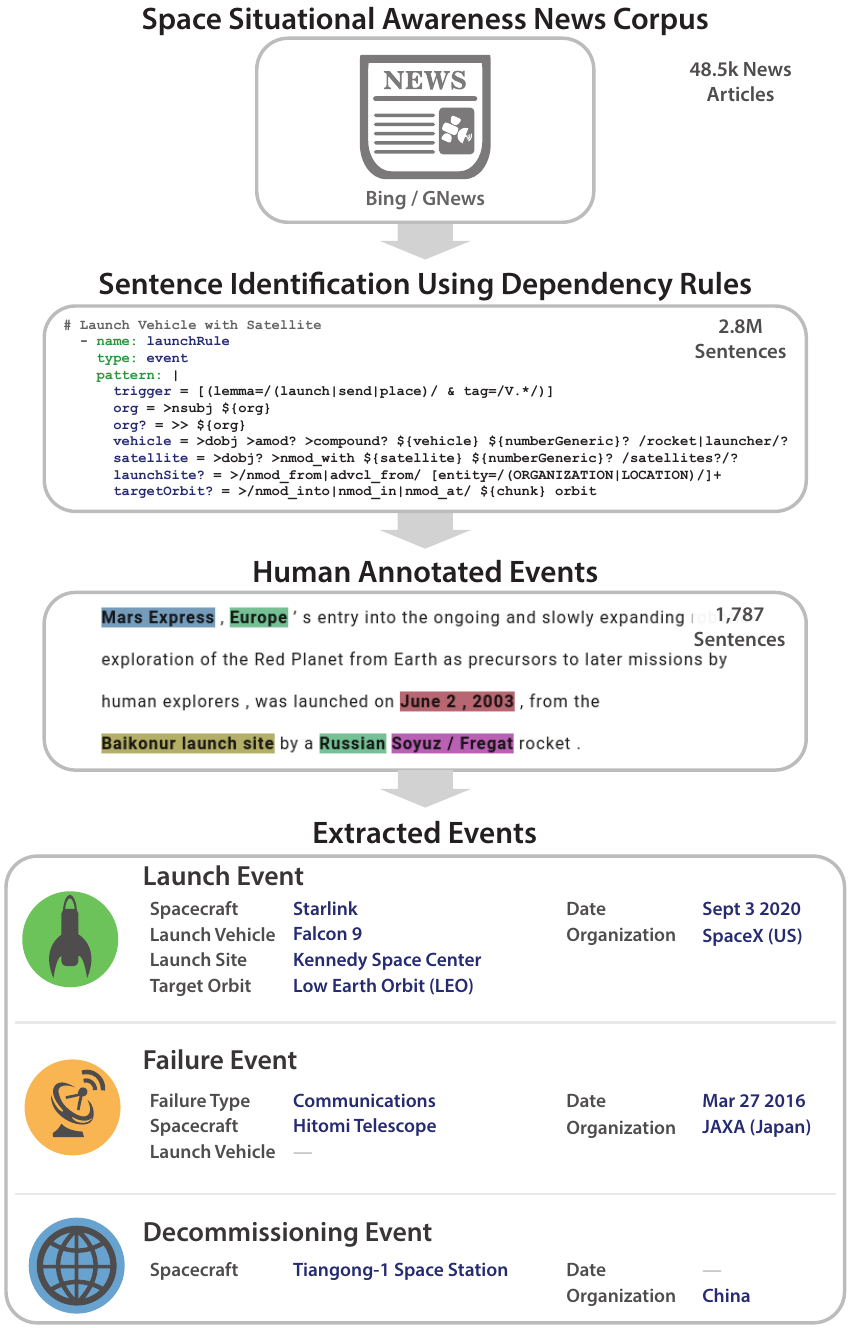}
    \caption{An overview of the proposed system.  A corpus of 48.5k space-domain news articles collected from the web serves as input to a sentence-level dependency-based information extraction system, \textsc{Odinson}, that isolates a shortlist of sentences for human annotation.  These are then used to train and evaluate a language-model based event extraction system, which extracts spacecraft \textit{launch, failure,} and \textit{decommissioning} events from text.}
    \label{fig:process}
\end{figure}

\indent In this work we explore extracting space-domain events from non-physical measurements, namely textual input from press releases and news articles. SSA is frequently accomplished by collecting and fusing multi-modal data \cite{esteva2020modeling,le2020representing}, where text is a comparatively inexpensive knowledge resource.  We hypothesize that text can supply information complementary to physics-based measurements, such as the published causes of spacecraft failures, whether spacecraft operators have planned astronavigation maneuvers, or whether a spacecraft has been safely decommissioned according to United Nations guidelines.  

\begin{figure*}[t!]
    \centering
    \includegraphics[width=6.25in]{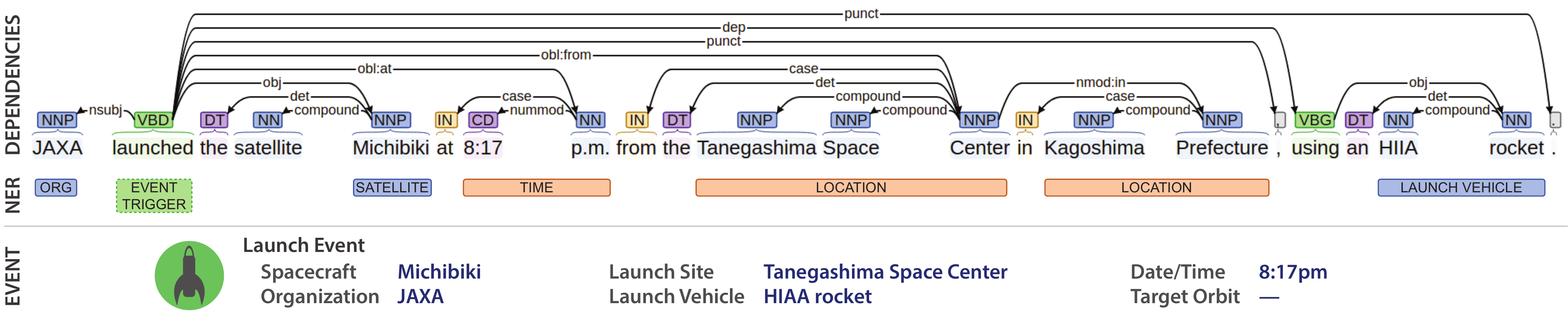}
    \caption{An example of constructing a dependency rule to find candidate \textit{launch} events for manual annotation. Starting with a trigger phrase (here, \textit{launched}), our system traverses specific syntactic dependencies most likely to contain information to fill specific event slots.  Both a space-domain \textit{(blue)} and generic CoreNLP \textit{(orange)} NER system assist in identifying slot fillers in high-confidence rules, while low-confidence rules fill slots by extracting chunks from likely dependencies. }
    \label{fig:dependencies}
\end{figure*}

Information extraction (IE) is often reported in high-resource domains where favorable performance is possible using pre-trained models.  In contrast, there exist high-importance domains (e.g. nuclear, defense, etc.) that remain under-explored due to low resource availability, and low task transfer from near-domains.  Space-domain information extraction poses these challenges in abundance -- comparatively few spacecraft have ever existed, and they are meticulously engineered, making launches, failures, and other SSA-relevant events infrequent but highly impactful. By automatically monitoring and extracting this information from sources available on the web, SSA users can augment existing physical measurements with behavioral information, or provide targets of interest for physical monitoring based on information gathered from text.



In this work we construct the first space-domain event corpus and extraction system, targeted at three demonstrative high-impact SSA events: spacecraft launches, failures, and decommissionings.  The contributions of this work are:

\begin{enumerate}
\item We collect a large-scale multi-year corpus of 48.5k news articles (containing 2.8M sentences) covering all 4,157 known satellites existing between 2009 through 2020.  We pair this with a dependency-rule based extractor that identifies 1,787 sentences with valid launch, failure, or decommissioning events, that we manually annotate with 15.9k labeled tokens representing event slots in each sentence.
\item We empirically demonstrate that in this low-resource high-impact domain, a state-of-the-art language-model-based extraction tool achieves F1 scores between 53 through 91 for extracting different event slots.  We highlight common causes of failure, including the large number of distractors in this corpus, in an error analysis.
\item We release our space-domain event corpus and extraction system as open source\footnote{Available at: \url{https://github.com/cognitiveailab/ssa-corpus}}. 
\end{enumerate}

\section{Approach}







An overview of our approach is shown in Figure~\ref{fig:process}. We approach this SSA task as a multi-slot information extraction problem.  Though event information may exist across sentences or entire documents, in this work we focus on events contained within single sentences.  
First, we assemble a large corpus of space-domain news articles.  We then use a dependency-based extraction framework to identify a shortlist of sentences that contain valid events, then manually annotate their event slots as spans in each sentence.  Finally, we evaluate the performance of a language-model-based extraction system at identifying these event spans.

\subsection{Multi-Slot Event Schema}
We focus on extracting three high-impact and comparatively frequent events in the corpus -- \textit{spacecraft launches, spacecraft failures, and spacecraft decommissionings}.  While IE is often framed as a triple extraction task \cite{etzioni2008open}, here each event schema has between 3 and 6 optional slots to populate with event-relevant information.  For example, the \textsc{Launch} event is used to describe the launch of a specific satellite, and may include additional information such as the launch vehicle used (e.g. an \textit{Atlas V}), the launch site (e.g. \textit{Cape Canaveral}), the target orbit (e.g. \textit{geosynchronous}), an organization the launch is on behalf of, and the date the launch occurred. A complete list of the event slots in each schema is included in Table~\ref{tab:performance}.

\begin{table*}[!t]
\centering
\small
\begin{tabular}{lcccccccccccc}
\toprule
		        & &  \multicolumn{3}{c}{Training} 	& & \multicolumn{3}{c}{Development}   & &   \multicolumn{3}{c}{Test}	\\\cmidrule(lr){3-5}\cmidrule(lr){7-9}\cmidrule(lr){11-13}
                & &        & Tokens &  Tokens   & &    & Tokens &  Tokens   & &    & Tokens &  Tokens   \\
SSA Event       & & Sents. & \textit{(w/tag)} & \textit{(total)} & & Sents.  & \textit{(w/tag)} & \textit{(total)} & & Sents. & \textit{(w/tag)} & \textit{(total)}  \\ 
\midrule
Launch		       & &  537  & 5,646 &  25,855   & &   350 & 2,890 & 13,357    & &   350 & 3,059 & 12,747 \\
Failure		       & &  310  & 2,748 &  12,905   & &   63 & 580 & 2,656        & &   63 & 449 & 2,043 \\
Decommissioning	   & &  81  & 396 &  3,064       & &   17 & 75 & 487           & &   17 & 68 & 504 \\
\bottomrule
\end{tabular}
\caption{\label{tab:summarystatistics} Overall statistics for the space situational awareness event corpus across the three events.  Sentences represents the number of sentences within a given set.  Tagged tokens represents the number of \textit{BIO-tagged} tokens that have a either a \textit{Beginning (B)} or \textit{Inside (I)} tag. }
\end{table*}

\section{Corpus and Annotation}

\subsection{Space Domain News Article Corpus}
We assembled a list of all known unclassified satellites active at any time between 2009 through 2020 using archived versions of the publicly available UCS satellite database\footnote{\url{https://www.ucsusa.org/resources/satellite-database}}, finding a total of 4,157 satellites over this 11 year period.  We collected a corpus of space-domain news articles by querying the Microsoft Bing News\footnote{\url{http://azure.microsoft.com}} and GNews\footnote{\url{http://gnews.io}} APIs for each satellite name (and any alternate names listed by USC), and retrieving the top 100 news articles for each name.  As news APIs regularly restrict results to a short temporal period (typically 30 days), and satellite events happen infrequently, collection took place over several years.  In total, collection returned 49,908 articles, which were post-processed with Boilerpipe \cite{kohlschutter2010boilerplate} to extract article content while removing advertisements and other extraneous information.  News organizations regularly publish syndicated duplicate articles, or republish updated articles months or years later.  To prevent leakage between training and unseen sets, any articles whose unigram cosine similarities exceed a threshold of 0.90 were pooled into the same set. 
The corpus was divided into training (28.7k document), and unseen(19.8k document) sets, with the most-recently collected articles and newest spacecraft present in the unseen set. The unseen set was evenly split into development and test sets after additional manual filtering described below.








%

\subsection{Identifying Sentences for Human Labelling with Dependency Rules}

The space-domain news article corpus contains 2.8M sentences across 48.5k documents.  To identify a shortlist of candidate sentences for human labeling that describe space events of interest (launches, failures, and decommissionings), we constructed a set of high-recall dependency-based extraction rules using the \textsc{Odinson} Information Extraction language \cite{valenzuela2020odinson}. \textsc{Odinson} allows expressing extraction rules as a combination of syntactic dependencies, named entities, part-of-speech tags, and surface forms, and processes documents approximately 5 orders of magnitude faster than comparable approaches \cite{wang2018scalable,valenzuela2016odin}, scanning and extracting sentences in our nearly 50k document corpus in approximately 4 minutes on inexpensive desktop hardware.  Articles were indexed using the \textsc{Odinson} indexer, where Stanford CoreNLP \cite{manning2014stanford} supplied tagging, lemmatization, and dependency parsing.


%
%
\begin{table*}[!t]
\centering
\small
\begin{tabular}{llccccccccccccl}
\toprule
    & 		                    & & 	\multicolumn{4}{c}{Development}	& &  \multicolumn{4}{c}{Test}	\\\cmidrule(lr){4-7}\cmidrule(lr){9-12}
 \multicolumn{2}{l}{Event and Slot}	& & 	    Pr	    & Re	    & F1    &   N    & &	    Pr	& Re & F1 & N  & &    Example	\\
\midrule
	
 \multicolumn{6}{l}{\textit{Launch Event}}	   & \textit{350} & & & & & \textit{349} \\
 ~~   & SatelliteName             & ~~ &      83      &   89      &   86  &     540   & ~~ &     85      &   89     &   87 &     528     & ~~ & Hubble Space Telescope \\
    & LaunchVehicle             & &      89      &   89       &   89  &     141   & &     86      &   92      &   89 &     170       & &  Space Shuttle STS-32 \\
    & LaunchSite                & &      86     &   97       &   91  &     88   & &     89     &   88      &   88 &     137       & &  Cape Canaveral \\ 
    & TargetOrbit               & &      62      &   84       &   71  &     25   & &     89     &   82      &   74 &     38       & &  Low-Earth Orbit \\
    
~\\
 \multicolumn{6}{l}{\textit{Failure Event}}	   & \textit{63} & & & & & \textit{63} \\
    & SatelliteName            & &      76      &   87      &   81   &     70   & &     69       &   87      &   77 &     53       & &  Telkom-3    \\
    & LaunchVehicle            & &      80      &   84      &   82   &     38   & &     88       &   94      &   91 &     31       & &  Proton-M    \\
    & FailureType              & &      41      &   55      &   47   &     65   & &     48      &   66     &   56  &    50        & &  Launch or power failure    \\
    
~\\
 \multicolumn{6}{l}{\textit{Decommissioning Event}}	   & \textit{17} & & & & & \textit{17} \\
    & SatelliteName            & &      61      &   85     &   71   &     20   & &     44       &   67     &   53 &     18       & &  NOAA-19    \\

~\\
 \multicolumn{10}{l}{\textit{Generic Slots Across Events}}\\
    & Organization             & &      66     &   81      &   73   &     349    & &     71      &   82      &   76  &     375       & &    NASA \\
    & Date                     & &      83     &   91      &   87   &     434    & &     84      &   88      &   86  &     390        & &    April 4, 1990 \\ [1mm]
\bottomrule
\end{tabular}
\caption{\label{tab:performance} Overall performance for the language model-based extraction system over the three event schema (\textit{spacecraft launches, failures, and decommissionings}), broken down by event slot.  Generic slots can optionally occur across all three event schema, and their performance is reported as the micro-average across all events.  \textit{N} for events represents the number of sentences in a given set (development, test) for that event, while for slots \textit{N} represents to the number of tagged spans across all sentences in a given set. Sentences occasionally contain more than one event, and as such the number of slots (e.g. SatelliteName) can exceed the number of sentences. }
\vspace{-2mm}
\end{table*}

\subsubsection{Named Entity Recognition}
To support rule-based extraction with \textsc{Odinson}, we constructed a named entity recognition system for domain-specific slot fillers, including \textsc{Spacecraft} (such as \textit{Hubble Space Telescope}), \textsc{LaunchVehicles} (such as \textit{Falcon 9}), and \textsc{LaunchSites} (such as \textit{Baikonur Cosmodrome}).  For open-domain entities, we use CoreNLP entities directly (e.g. \textsc{Date}), or a combination of both CoreNLP and our space-domain NER system (e.g. \textsc{Organization}).

\subsubsection{Rule Authoring}
Rules were constructed through an interative process of querying the training set for high-frequency n-grams that might serve as \textsc{Odinson} trigger phrases associated with each of the 3 events (e.g. \textit{launched, failure,} or \textit{decommissioned}), then authoring slot-extraction rule components based on traversing key syntactic dependencies attached to the trigger phrase.  An example dependency tree and associated event is shown in Figure~\ref{fig:dependencies}.  To balance precision and robustness, we constructed 35 high-confidence rules that required NER matches for all slots, while 32 lower-confidence back-off rules allow matching unknown entities or generic NP-chunks for one or more slots, increasing recall for (for example) newer or unknown spacecraft at the expense of precision.  In total, 67 rules were authored across all 3 events. 

\subsubsection{Manual Filtering}
The rule-based system identified 4346 \textit{launch}, 824 \textit{failure}, and 218 \textit{decommissioning} candidate sentences from the corpus of 2.8M sentences.  Manual inspection showed that 91\% of launch sentences and 53\% of both failure and decommisioning sentences contained a valid event, while others contained unrelated information (e.g. the launch of a new product) and were filtered out.  Due to the large number of events, the \textit{launch} set was randomly subsampled to 30\% of its original size for human labeling.  A total of 1,787 sentences progressed through this stage, for subsequent labeling by human annotators. 

\subsection{Annotating Space Events}

Annotation took the form of a span-labelling task, where for a given sentence an annotator would highlight relevant spans of text (e.g. \textit{``Hubble Space Telescope''}), and select an appropriate event slot label (e.g. \textsc{Spacecraft}). A single sentence can mention more than one event (for example, two satellites being launched), and in these cases all relevant slot fillers are annotated for both events.\footnote{Sentences without mention of a specific \textsc{Spacecraft} (for launches and decommissionings) or a specific \textsc{Spacecraft} or \textsc{LaunchVehicle} (for failures) were removed from consideration during the manual filtering step.}  We used LightTag\footnote{\url{www.lighttag.io}} \cite{perry-2021-lighttag} for annotation and calculating interannotator agreement statistics. 

Four annotators completed the labeling task, with each sentence being independently labelled by three annotators.  After this labelling, the annotators then used the review functionality of LightTag to discuss and resolve disagreements.  Individual agreement relative to this final consensus ranged between 80\% to 90\% precision and 78\% to 87\% recall across the four annotators.  

Summary statistics of the labelled sets are shown in Table~\ref{tab:summarystatistics}.  The final corpus contains a total of 73.6k tokens across 1,787 sentences, with 15.9k tokens (22\%) containing an event slot label.

\section{Modelling}








\subsection{Multi-slot Model}
Recently a number of off-the-shelf methods for state-of-the-art triple extraction have emerged \cite{han-etal-2019-opennre}, but few of these systems work in multi-slot scenarios. Several groups \cite{zhang2020rapid,liu2019open} demonstrate multi-slot extraction using pre-trained embedding models such as BERT \cite{devlin2019bert}, but their systems are not publicly available.  We recreate a similar system by framing the multi-slot extraction problem as a sequence labeling task where entity spans are labeled with a given slot name (e.g. \textsc{SatelliteName}, \textsc{LaunchVehicle}, etc.), and adapt an off-the-shelf BERT-based sequence labeling system\footnote{\url{https://github.com/kamalkraj/BERT-NER}} to this task, which achieves near state-of-the-art performance on entity labeling tasks \cite{smith2020scienceexamcer}.

\subsection{Results}

Each of the three events was trained and evaluated separately.  We made use of the \texttt{BERT-Base-cased} (110M parameter) model.  Performance was tuned on the development set, where we observed that performance peaked at 60 training epochs.  We report performance using the standard definitions of precision, recall, and F1 \cite[inter alia]{manning1999foundations}.

Overall extraction performance for each SSA event is shown in Table~\ref{tab:performance}.  Overall performance per slot ranges from 47 to 91 F1, with higher-frequency slots that tend to follow regular patterns tending to perform better than low-frequency categories with less regular patterns.  \textsc{SatelliteName}, \textsc{LaunchVehicle}, and \textsc{LaunchSite} were generally able to achieve fair to excellent performance (peaking near 90 F1 for the \textit{Launch} event), while slots with comparatively less training data and higher variation in their presentation in sentences (e.g. \textsc{TargetOrbit} and \textsc{FailureType}) achieved modest performance approaching 74 and 56 F1 on the test set, respectively.  \textsc{FailureType} is the lowest-performing slot, highlighting the challenge of identifying spans of text that describe failures when they may present as either relatively frequent high-level causes (e.g. \textit{``launch failure''} or \textit{``power failure''}), or more-specific descriptions less-frequently observed during training (e.g. \textit{``fuel leak''}, or \textit{``problems with antenna''}) that more precisely convey critical issues.

\begin{table}[!t]
\centering
\begin{tabular}{cl}
Prop.  & Error	\\
\hline
43\%		& Mention not relevant to event	\\
21\%		& Span errors (too long or short)\\
8\%		    & Inferred label is possibly relevant	\\
7\%	    	& Gold label is incorrect	\\
5\%         & Abbreviations mistaken as satellite name \\
\hline
\end{tabular}
\caption{\label{tab:erroranalysis} Common categories of errors on a cross-section of 100 randomly selected errors on the test set.  Proportions do not sum to 100\% as less frequent error classes are omitted.}
\end{table}

\subsection{Error Analysis}

To better understand the challenges this dataset poses to extraction systems, we analyzed 100 randomly-selected extraction errors on the test set and identified five major classes of error, outlined in Table~\ref{tab:erroranalysis}, and described below:

{\flushleft\textbf{Mention Not Relevant (43\%):}} Sentences containing space events are frequently long and contain historical or other contextual information that serve as distractors for the events being extracted.  For example, in the \textsc{Failure} sentence \textit{``AMC-14 was delayed to February due to the failure of a Proton [LaunchVehicle] rocket in September [Date]''}, the model populates the \textsc{SatelliteName} slot with AMC-14 even though it is not involved in the failure event. 

{\flushleft\textbf{Span Errors (21\%):}} Span errors are when the model captures part of a mention, but does not overlap completely with the gold information -- typically from ending too early, starting too late, or missing tags in the middle of a span (for example, missing \textit{``of''} in \textit{``United States of America''}. 

{\flushleft\textbf{Possibly Relevant (8\%):}} While not identical to gold spans, in 8\% of cases, the spans chosen by the model could also be considered valid by manual judgement.  For example, in one error, the model broke the gold \textsc{TargetOrbit} span \textit{``highly elliptical and highly inclined orbit''} into two separate mentions \textit{``highly elliptical''} and \textit{``highly inclined orbit''}, that are both valid. 

{\flushleft\textbf{Errors in gold labels (7\%):}} In 7\% of errors, the gold annotation had technical errors.  For example, sentences automatically extracted from news articles occasionally contain extraneous out-of-sentence information (such as part of the headline, or captions from photographs) that might contain the same entities as those in the event, and our annotation protocol specifies these extraneous spans not to be annotated -- but sometimes the borders are difficult to determine.  With the overall micro-precision of the model at 80\% across test set events, and 7\% of model errors due to annotation errors, we estimate the total accuracy of this annotation after the review procedure to be approximately 98\%.\footnote{Approximately 20\% of tags predicted by the model were errorful on the test set.  With a manually estimated 7\% error rate in gold labels on these errors, and assuming correctly predicted gold labels are correct labels, we estimate the overall error rate in annotating gold labels to be approximately 2\%.} 

{\flushleft\textbf{Abbreviations as Satellites (5\%):}} Abbreviations in text (e.g. \textit{``AFP''}, a press organization) are occasionally mistaken as \textsc{SatelliteName} slots due to the model learning that satellite names are frequently capitalized.


%
%

\section{Conclusion}





We present the first Space Situational Awareness event corpus and extraction system that can monitor news articles and extract three high-impact events: spacecraft launches, failures, and decommissionings.  The corpus contains 1,787 labelled sentences with 15.9k manually labelled tokens, drawn from a corpus of nearly 48.5k news articles spanning all 4,157 known satellites active between 2009 and 2020. Our analysis shows that baseline model performance (F1) ranges between 53 and 91 per event slot, highlighting the challenges associated with this low-resource domain.  The corpus and extraction system are open source, available at: \url{https://github.com/cognitiveailab/ssa-corpus} . 



\section{Bibliographical References}\label{reference}

\bibliographystyle{lrec2022-bib}
\bibliography{lrec2022-example}


\end{document}